\documentclass[runningheads]{llncs}
\usepackage[T1]{fontenc}
\usepackage{graphicx}
\usepackage[compatibility=false]{caption}
\usepackage{subcaption}
\usepackage{booktabs}
\usepackage{hyperref}
\usepackage{subcaption}
\usepackage{multirow}
\usepackage{xcolor}
\usepackage{float}
\usepackage{cite}

\begin{document}
\title{YOLO11-4K: An Efficient Architecture for Real-Time Small Object Detection in 4K Panoramic Images}
\titlerunning{YOLO11-4K}
%
\author{Huma Hafeez\inst{1}\orcidID{0009-0005-0290-3531} \and
Matthew Garratt\inst{1}\orcidID{0000-0003-0222-430X} \and
Jo Plested\inst{2}\orcidID{0000-0001-9342-4539}
\and
Sankaran Iyer\inst{3}\orcidID{0000-0001-7580-5602}
\and
Arcot Sowmya\inst{3}\orcidID{0000-0001-9236-5063}}
\authorrunning{H. Hafeez et al.}
%
\institute{School of Engineering \& Technology, University of New South Wales, Canberra, Australia\\ \and
School of Systems \& Computing, University of New South Wales, Canberra, Australia\\
\and
School of Computer Science \& Engineering, University of New South Wales, Sydney, Australia\\
}
\maketitle              
\begin{abstract}
The processing of omni-directional 360$^{\circ}$ images poses significant challenges for object detection due to inherent spatial distortions, wide fields of view, and ultra-high-resolution inputs. Conventional detectors such as YOLO are optimised for standard image sizes (e.g., 640×640 pixels) and often struggle with the computational demands of 4K or higher-resolution imagery typical of 360$^{\circ}$ vision. To address these limitations, we introduce YOLO11-4K, an efficient real-time detection framework tailored for 4K panoramic images. The architecture incorporates a novel multi-scale detection head with a P2 layer to improve sensitivity to small objects often missed at coarser scales, and a GhostConv-based backbone to reduce computational complexity without sacrificing representational power. To enable evaluation, we manually annotated the CVIP360 dataset, generating 6,876 frame-level bounding boxes and producing a publicly available, detection-ready benchmark for 4K panoramic scenes. YOLO11-4K achieves 0.95 mAP@50 with 28.3 ms inference per frame, representing a 75\% latency reduction compared to YOLO11 (112.3 ms), while also improving accuracy (mAP@50 = 0.95 vs. 0.908). This balance of efficiency and precision enables robust object detection in expansive 360$^{\circ}$ environments, making the framework highly suitable for real-world high-resolution Panoramic applications. Whilst this work focuses on 4K omnidirectional images, the approach is broadly applicable to high-resolution detection tasks in domains such as autonomous navigation, surveillance and augmented reality.

\end{abstract}

\section{Introduction}
\label{sec:intro}
Real-time object detection has advanced significantly with the development of one-stage detectors, especially the YOLO series ~\cite{redmon2016you,redmon2017yolo9000,redmon2018yolov3,bochkovskiy2020yolov4,jocher2020ultralytics,li2022yolov6,wang2023yolov7,sohan2024review,wang2024yolov9,wang2024yolov10,yolo11_ultralytics}. These architectures effectively balance speed and accuracy for medium-resolution images, typically around 640×640 pixels. However, scaling these detectors to ultra-high-resolution inputs such as 4K (3840×2160 pixels) is challenging due to the large computational cost and slower inference times, which especially affect the detection of small objects. These challenges are even greater in Panoramic vision systems, where equirectangular projection causes geometric distortions and non-uniform scaling of objects across the 360$^{\circ}$ image. Maintaining native 4K resolution is important for detecting small and distant objects accurately, however doing so in real time is challenging.

Previous efforts, such as R\r{u}\v{z}i\v{c}ka and Franchetti's two-stage attention pipeline \cite{ruuzivcka2018fast}, attempted to address high-resolution detection by applying coarse-to-fine inference using YOLOv2. Their method leverages GPU parallelism to improve computational efficiency and achieves 3–6 frames per second (FPS) for 4K and 2 FPS for 8K video. While effective, such multi-stage strategies increase complexity and are not optimised for end-to-end real-time inference. To address these limitations, we propose YOLO11-4K, a fully convolutional detection framework tailored for real-time performance on 4K Panoramic images. Our contributions include:

\begin{enumerate}
    \item End-to-end 4K Panoramic detection: YOLO11-4K directly processes full-resolution images without downsampling or cropping, preserving critical spatial information for detection across the entire 360$^{\circ}$ field. 
    
    \item We manually annotated the CVIP360 dataset, producing 6,876 frame-level bounding boxes to create 
 detection-ready benchmark for 4K panoramic scenes, enabling rigorous evaluation of high-resolution small object detection. The annotated dataset has already been published on Github.
    
    \item Improved small object detection: We add a P2 detection head (an additional early prediction layer operating on high-resolution feature maps) early in the network to capture finer spatial details, improving detection of small objects that are often missed by standard detection layers.

    \item Efficient computation: Using lightweight convolutional modules such as GhostConv for generating feature maps with fewer parameters, C3Ghost for efficient feature aggregation and C3k2 for reduced kernel complexity, we reduce computations  while maintaining strong feature representation and detection accuracy.
\end{enumerate}

With these improvements, YOLO11-4K achieves real-time inference at 21.4 ms per 4K frame on modern GPUs, substantially outperforming standard YOLO architectures  while preserving detection accuracy. This work presents a scalable and practical solution for small object detection in high-resolution 360$^{\circ}$ scenes, with potential applications in autonomous systems, surveillance and augmented reality.

\section{YOLO11-4K Architecture}
\label{sec:architecture}

YOLO11-4K is a specialised extension of the Ultralytics YOLO11 \cite{yolo11_ultralytics} framework, and is designed to address the challenges of real-time small object detection in ultra-high-resolution Panoramic images (3840×3840 pixels). These images introduce severe computational and geometric complexities due to their scale and the non-uniform spatial distribution of objects. To overcome these issues, YOLO11-4K integrates several architectural modifications including a P2 detection layer for fine-scale feature extraction, lightweight GhostConv and C3Ghost modules for efficient representation learning, and end-to-end full-resolution processing, to enable accurate real-time detection across the entire 360$^\circ$ field of view.

\subsection{Design Overview}
The YOLO11-4K architecture, illustrated in Fig. ~\ref{architecture}, comprises three key components:
\begin{enumerate}
    \item Explicit support for 4K equirectangular panoramic (ERP) input, enabling precise detection in wide 360$^\circ$ fields of view while preserving projection-
\setlength{\fboxsep}{5pt}
\begin{figure}[H]
    \centering
    \fbox{%
    \includegraphics[width=\linewidth]{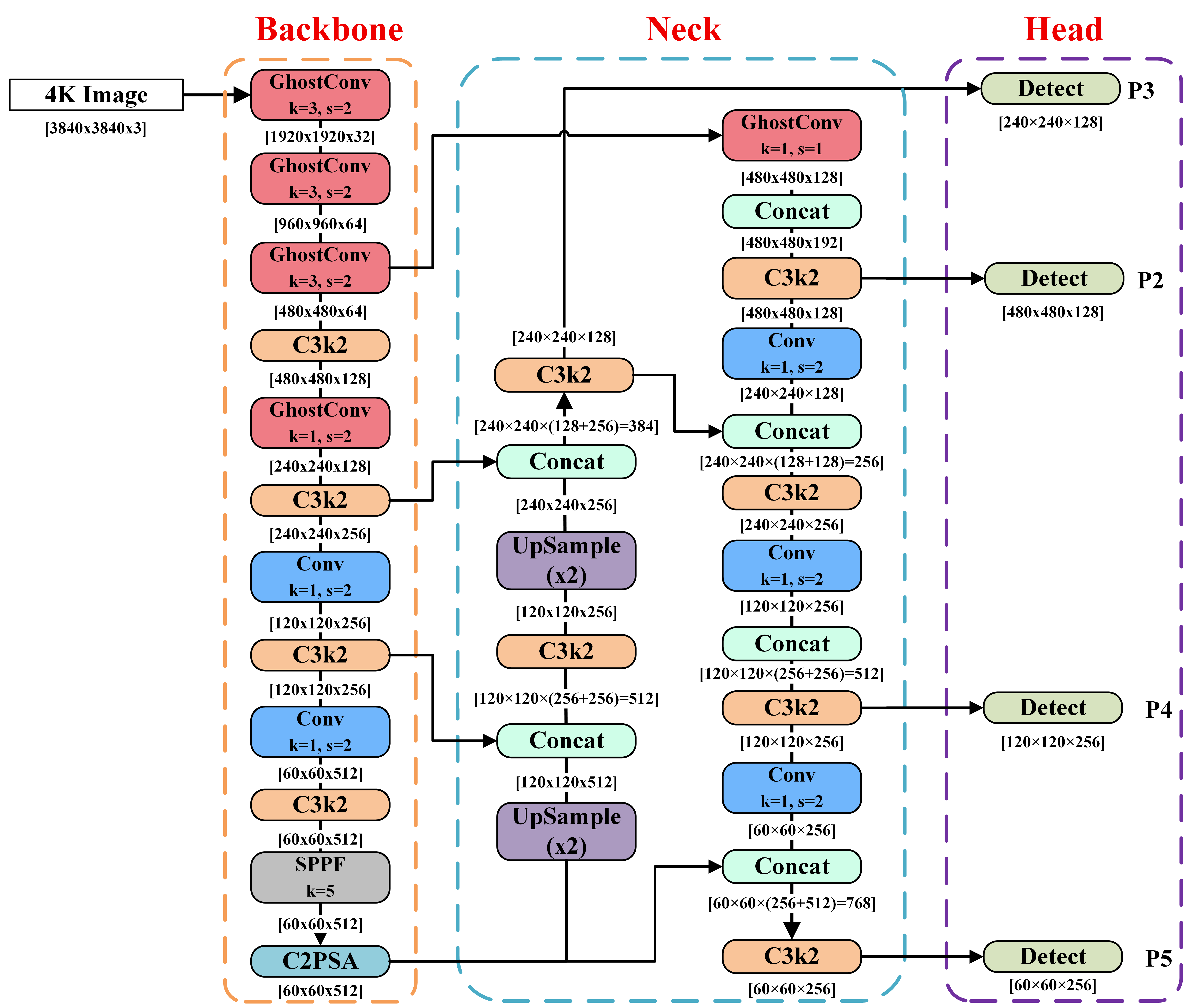}%
        }
    \caption{YOLO11-4K architecture supporting end-to-end inference on 4K ERP inputs for full 360$^\circ$ spatial coverage}
    \label{architecture}
\end{figure}

     related spatial distortions,
    \item A GhostConv based backbone designed for efficient, low-cost feature extraction, and
    \item A four-scale detection head (P2–P5) utilising C3k2 blocks to improve the detection of both small and large objects. 
\end{enumerate}
This architectural design demonstrates that real-time processing can be achieved alongside ultra-high-resolution input handling. By carefully selecting lightweight convolutional modules and strategically extending the detection head, YOLO11-4K offers a practical and efficient solution for high-resolution object detection in 360$^\circ$  environments.

\subsection{Support for High-Resolution Input}
A defining feature of YOLO11-4K is its ability to directly process 4K resolution inputs (3840×3840 pixels), a substantial increase compared to the standard 640×640 resolution used in prior YOLO models. This high-resolution processing is essential for preserving fine spatial details, particularly for small and distant objects distributed across the 360$^\circ$ field of view in Panoramic images. Although the increase in input resolution imposes greater computational demands, YOLO11-4K offsets this overhead through the use of lightweight GhostConv modules and efficient backbone design, enabling real-time inference without compromising detection accuracy.

\subsection{Lightweight Backbone}
The YOLO11-4K backbone is designed to extract hierarchical features efficiently from high-resolution Panoramic inputs. To mitigate redundancy in standard convolutions, the detector employs ghost convolution \cite{huang2024improvement}, which improve efficiency 
by generating many feature maps using cheap linear operations instead of expensive convolutions, reducing redundancy and computation while maintaining accuracy. The operation first applies a 1×1 convolution to capture intrinsic features, then uses low-cost 5×5 linear transformations to derive additional maps. These are concatenated to form the final output, an essential consideration for processing 3840×3840 pixel images in real time.

Subsequent stages integrate several core components from the YOLO11 architecture \cite{khanam2024yolov11}. These include the C3k2 module, a computationally efficient variant of the standard C3 block achieved by substituting 3×3 bottleneck convolutions with 2×2 kernels; the Spatial Pyramid Pooling–Fast (SPPF) module, which captures multi-scale context through sequential max-pooling with minimal overhead; and the C2PSA attention mechanism, which adaptively enhances spatial and channel-wise feature responses to improve representational fidelity.

\subsection{Small Object Aware Detection Head-P2}
In the original YOLO11 design, detection heads commonly operate on P3 (fine-grained, high-resolution spatial details suitable for small objects), P4 (balances detail and semantic context) and P5 (encodes more abstract, high-level features for larger objects) feature maps, which suffice for many applications. However, they may fail to capture the fine detail required for small object detection in very high-resolution images. To overcome this limitation, YOLO11-4K introduces a new P2 detection branch designed specifically for high-resolution small object detection. This branch extracts the early high-resolution feature map at 1/4 input scale directly from the backbone (tapping into the GhostConv output at layer 1). It then refines these features through a sequence of GhostConv and C3k2 modules.

The detection head in YOLO11-4K performs multi-scale fusion by progressively upsampling and concatenating feature maps across scales. Starting from the deepest P5 features, the model upsamples and concatenates them with P4 features, refining the combined output with C3k2 modules. This process is repeated to fuse P3 features, enabling detection across multiple scales. Parallel to this, the newly added P2 branch provides refined high-resolution features directly to the detection layer. Finally, the detection module aggregates outputs from P2, P3, P4 and P5 branches, allowing the network to simultaneously detect small, medium and large objects with improved accuracy.

By integrating GhostConv for computational efficiency, adding the P2 detection branch for enhanced small object sensitivity and scaling to 4K input resolution, YOLO11-4K represents a significant advancement in real-time object detection for high-resolution Panoramic imagery. The architecture maintains a careful balance between speed and precision, addressing the challenges inherent in small object detection within expansive visual fields.

\section{Implementation}
\label{sec:implementation}
The YOLO11-4K architecture was implemented by extending the Ultralytics YOLOv11 framework using the PyTorch deep learning library. Training was performed on high-resolution panoramic image datasets, leveraging mixed precision training to accelerate convergence and reduce GPU memory consumption. We employed optimised training schedules and data augmentation techniques to achieve a balance between detection accuracy and inference speed. The following subsections provide detailed descriptions of the training setup, datasets and implementation details.

\subsection{Training Configuration}
YOLO11-4K was trained using the Ultralytics training pipeline with a custom configuration tailored for 4K resolution inputs. The training process was run for 50 epochs, using an input image size of 3840×3840 pixels, and Automatic Mixed Precision training (amp=True) enabled, which leverages 16-bit computations for most operations while preserving 32-bit precision where necessary to reduce GPU memory usage and speed up training. To prevent overfitting, early stopping was implemented with a patience of 10 epochs. Due to the substantial memory demands of processing high-resolution images, the batch size was limited to one.

The training was conducted on the Australian National Computational Infrastructure (NCI) Gadi supercomputer, utilising the \emph{gpuvolta} queue which provides access to nodes equipped with dual NVIDIA Volta GPUs. This dual-GPU configuration was leveraged to enable efficient parallelism, thereby improving training throughput and reducing convergence time.

The training pipeline utilised the Ultralytics built-in hyperparameter tuning mechanism, which selected the AdamW optimiser with a learning rate of 0.001667 and $\beta_1 = 0.9$, $\beta_2 = 0.999$. The optimiser applied differentiated weight decay across parameter groups: 0.0005 for most weights, and 0.0 for others including biases.  Gradient accumulation (a technique where gradients are computed over several mini-batches before performing a weight update, effectively simulating a larger batch size when memory is limited) was disabled due to the 2-GPU setup. Gradient synchronisation was handled automatically by the data-parallel training logic in the Ultralytics framework. Despite the small per-GPU batch size, this configuration, combined with AdamW and appropriate learning rate scheduling, allowed successful convergence while preserving generalisation on high-resolution inputs.

\subsection{Dataset}
To train and evaluate the YOLO11-4K model on high-resolution panoramic scenes, we utilised the CVIP360 dataset \cite{mazzola2021dataset}. This dataset is specifically designed for 360$^\circ$ vision applications and comprises a diverse collection of high-resolution equirectangular projection (ERP) images captured using the GarminVIRB 360 camera, covering both indoor and outdoor scenes. Although originally released with depth estimation annotations, CVIP360 lacked the bounding box labels required for object detection. To overcome this limitation, we re-annotated the dataset using the Roboflow~\cite{roboflow} platform, extending CVIP360 into a detection-ready benchmark, enabling the training and evaluation of small object detectors in 360$^\circ$ environments.

\begin{figure}[h]
    \centering
    \fbox{%
    \includegraphics[width=\linewidth]{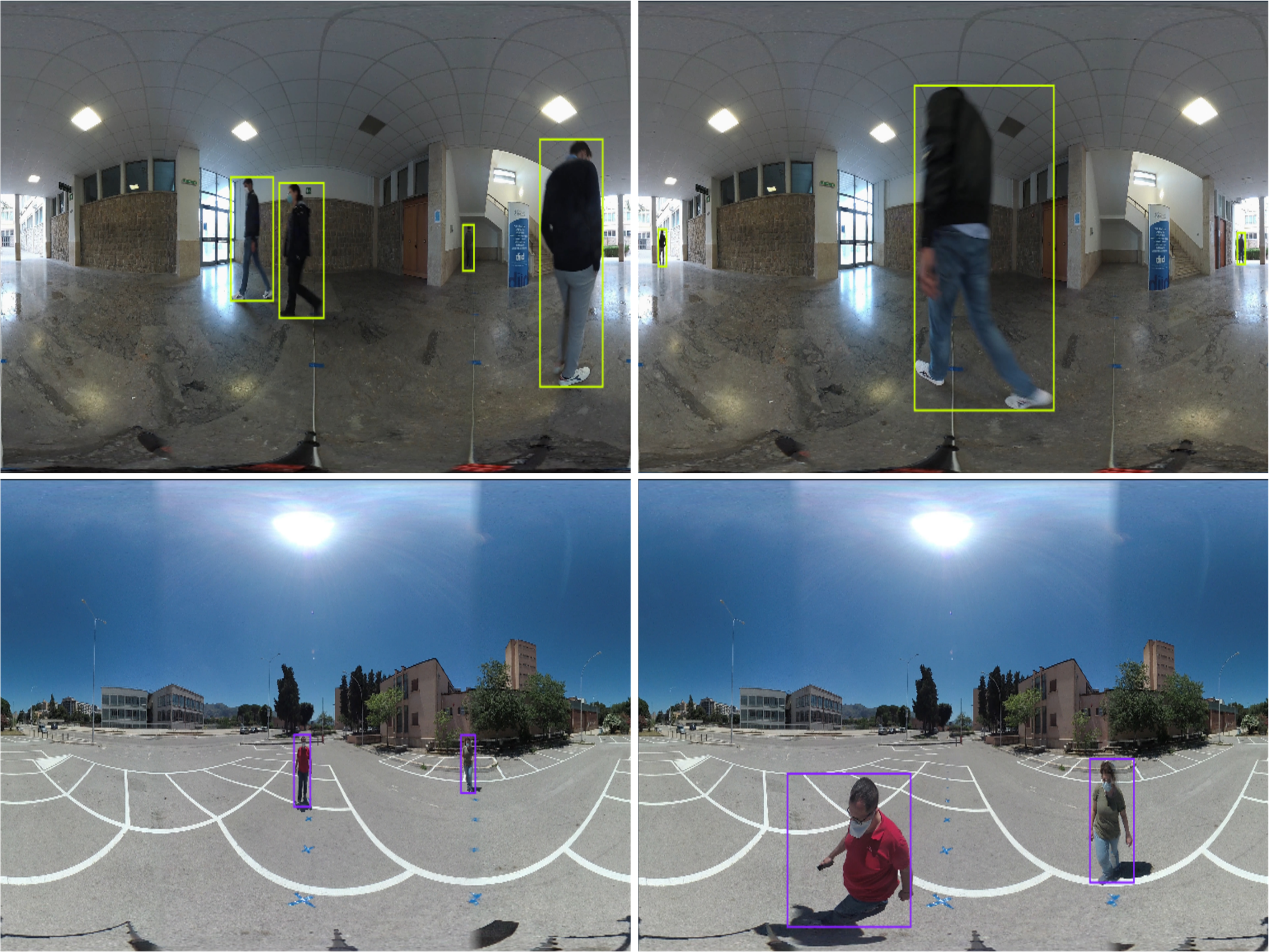}%
    }
    \caption{4K CVIP360 ERP images with annotated small and distant pedestrians in indoor and outdoor 360$^\circ$ scenes}
    \label{dataset}
\end{figure}

All images in the dataset are available in 4K resolution as shown in Fig \ref{dataset}, making them particularly suitable for small object detection and evaluation under wide field-of-view (FoV) conditions. The dataset primarily focuses on pedestrian detection within 360$^\circ$ video frames.  As the dataset was originally intended for depth estimation, it provided a single annotation file per video sequence. In order to facilitate frame-wise detection tasks, the dataset was manually re-annotated using the Roboflow platform, resulting in a total of 6,876 annotated images. These detection-ready annotations are publicly available on GitHub at:
\url{https://github.com/huma-96/CVIP360_BBox_Annotations}. To ensure robustness and mitigate bias due to dataset partitioning, we employed a 5-fold cross-validation strategy. The entire dataset was randomly divided into five equally sized folds. In each iteration, four folds were used for training and validation (with 80\% of these four folds for training and 20\% for validation), while the remaining fold served as a hold-out test set. Final performance metrics were obtained by averaging results across the five models.

\setlength{\fboxsep}{2pt}

Due to the geometric distortions present in ERP imagery, the CVIP360 dataset offers a challenging benchmark for evaluating model robustness in non-planar representations. The spatial distribution and scale variation of objects across the 360$^\circ$ FoV  align well with the objectives of this work, which aims to enhance detection performance within panoramic visual environments, especially for small and distant objects.

\subsubsection{Cross-dataset evaluation}
To evaluate cross-dataset generalisation beyond panoramic imagery, we additionally evaluated all models on the Multi-Resolution Traffic Monitoring Dataset (MRTMD) \cite{11071529} at 2160p. MRTMD is a multi-category traffic dataset containing classes such as bicycle, car, motorcycle and bus; it is characterised by dense scenes, extreme scale variation and many ultra-small distant objects. We use the same evaluation pipeline (detection thresholds and IoU settings) as for CVIP360 to ensure comparability.

\subsection{Evaluation Metrics}
To comprehensively evaluate the detection performance of YOLO11-4K, we adopted standard object detection metrics used in benchmark evaluations, with particular attention to real-time processing requirements and small object detection accuracy.

\textbf{Mean Average Precision (mAP):} mAP is the standard measure for object detection, computed by averaging the precision–recall performance (AP) across all classes, where AP corresponds to the area under the precision–recall curve for each class. To comprehensively assess detection accuracy, we report both mAP@0.50 and mAP@0.50:0.95. The mAP@0.50 score captures performance at a moderate Intersection over Union (IoU) threshold, indicating how well the model detects and classifies objects with at least 50\% overlap. In contrast, mAP@0.50:0.95 provides an averaged score over ten IoU thresholds (ranging from 0.50 to 0.95 with increments of 0.05), offering a stricter and more detailed evaluation of the model’s localisation precision and overall robustness.

\textbf{Precision and Recall:} We compute precision and recall across the validation and test sets to assess the model's ability to correctly identify true positives while minimising false positives and false negatives. These metrics are especially important for small object detection where class imbalance and localisation ambiguity are more common. Precision ($P$) and recall ($R$) are defined as:  
\begin{equation}
    P = \frac{TP}{TP + FP} 
    \label{Precision}
\end{equation}
\begin{equation}
    R = \frac{TP}{TP + FN} \label{Recall}
\end{equation}

where $TP$, $FP$ and $FN$ denote true positives, false positives and false negatives respectively.

\textbf{Inference Speed (FPS):} Real-time performance is critical for high-resolution panoramic applications. We measure the inference speed of YOLO11-4K in frames per second (FPS) using dual NVIDIA Volta GPUs and 4K (3840×3840) resolution inputs. This metric evaluates the practicality of deploying the model in time-sensitive scenarios.

By combining localisation precision, real-time throughput and scale-sensitive evaluation, this multi-metric framework enables a rigorous and holistic assessment of  model performance under the challenging conditions posed by high-resolution panoramic imagery.

\section{Results and Discussion}
\label{sec:results_and_discussion}
This section presents a detailed analysis of the performance of YOLO11-4K on the CVIP360 dataset, with comparisons across detection accuracy, speed and robustness for small object detection. We also discuss how the architectural choices, particularly the use of GhostConv, the inclusion of the P2 head and the use of high-resolution inputs, contributed to the observed improvements. In Table~\ref{CVIPresults} the performance achieved by YOLO11-4K, YOLO11 as well as some earlier versions of YOLO, on the validation and test splits of the CVIP360 dataset are summarised. 
\begin{table}[h]
\caption{Performance comparison of different detectors on the CVIP360 dataset}
\centering
\begin{tabular}{|l|p{1.5cm}|p{1.3cm}|p{1.2cm}|p{1.5cm}|p{1.5cm}|p{2cm}|}
\hline
\textbf{Model} & \textbf{Precision}& \textbf{\textit{Recall}}& \textbf{\textit{mAP@ 50}} & \textbf{\textit{mAP@ 50:95}} & \textbf{\textit{Inference (ms)}}& \textbf{Min. Object Size (px)}\\
\hline

YOLOv5 & 0.727 & 0.871 & 0.881 & 0.639 & 86.1&4.33x27.04\\

YOLOv8 & 0.779 & 0.863 & 0.898 & 0.662 & 88 &3.45x28.97\\

YOLO11 & 0.777 & 0.889 & 0.908 & 0.671 & 112.3&2.97x26.11\\

\textbf{YOLO11-4K} & \textbf{0.794} & \textbf{0.955} & \textbf{0.95} & \textbf{0.718} & \textbf{28.3}&\textbf{2.70x25.14}\\
\hline
\end{tabular}
\label{CVIPresults}
\end{table}

YOLO11-4K achieved the highest overall performance in precision, recall and mAP@50 while delivering the fastest inference among all evaluated models. A key advantage of YOLO11-4K is its inference efficiency. Processing 3840×3840 high-resolution panoramic images, it achieves an inference time of 21.4 ms per image, nearly five times as fast as YOLO11. This speedup is largely attributable to the hybrid backbone design, which combines GhostConv layers in shallow stages with standard convolutions in deeper layers and kernel-stride parameter tuning, efficiently balancing computational cost with representational capacity.

\begin{figure*}[t]
    \centering
    \fbox{%
    \includegraphics[width=\linewidth]{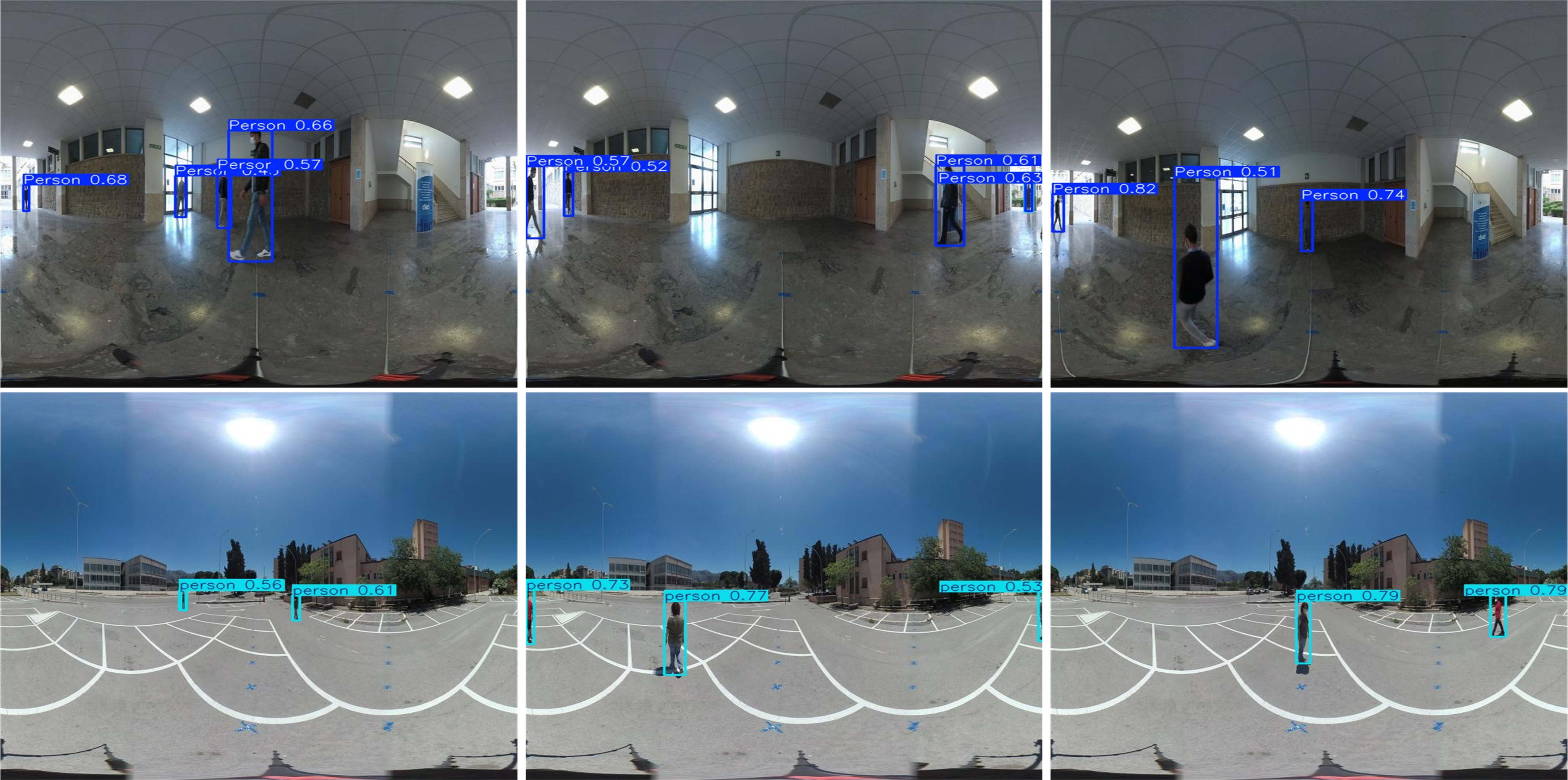}%
        }
    \caption{YOLO11-4K detection results on challenging 4K panoramic scenes. Both indoor and outdoor scenarios are shown, highlighting successful detection of small and occluded objects.}
    \label{results}
\end{figure*}

In Figure~\ref{results}, qualitative detection results of YOLO11-4K on both indoor and outdoor 4K panoramic scenes are illustrated. The model effectively detects small, medium and partially occluded objects across the entire 360$^{\circ}$ field of view. Across the test set, a total of 1,604 objects were detected, with average object dimensions of 28.9 × 133.2 pixels. The smallest object detected measured 2.7 × 25.1 pixels, while the largest measured 126.5 × 319.8 pixels. 
\setlength{\fboxsep}{5pt}

\setlength{\fboxsep}{5pt}

\begin{figure}[H]
    \centering
    \fbox{%
    \begin{minipage}{0.49\linewidth}
        \begin{subfigure}{\linewidth}
            \centering
            \includegraphics[width=\linewidth]{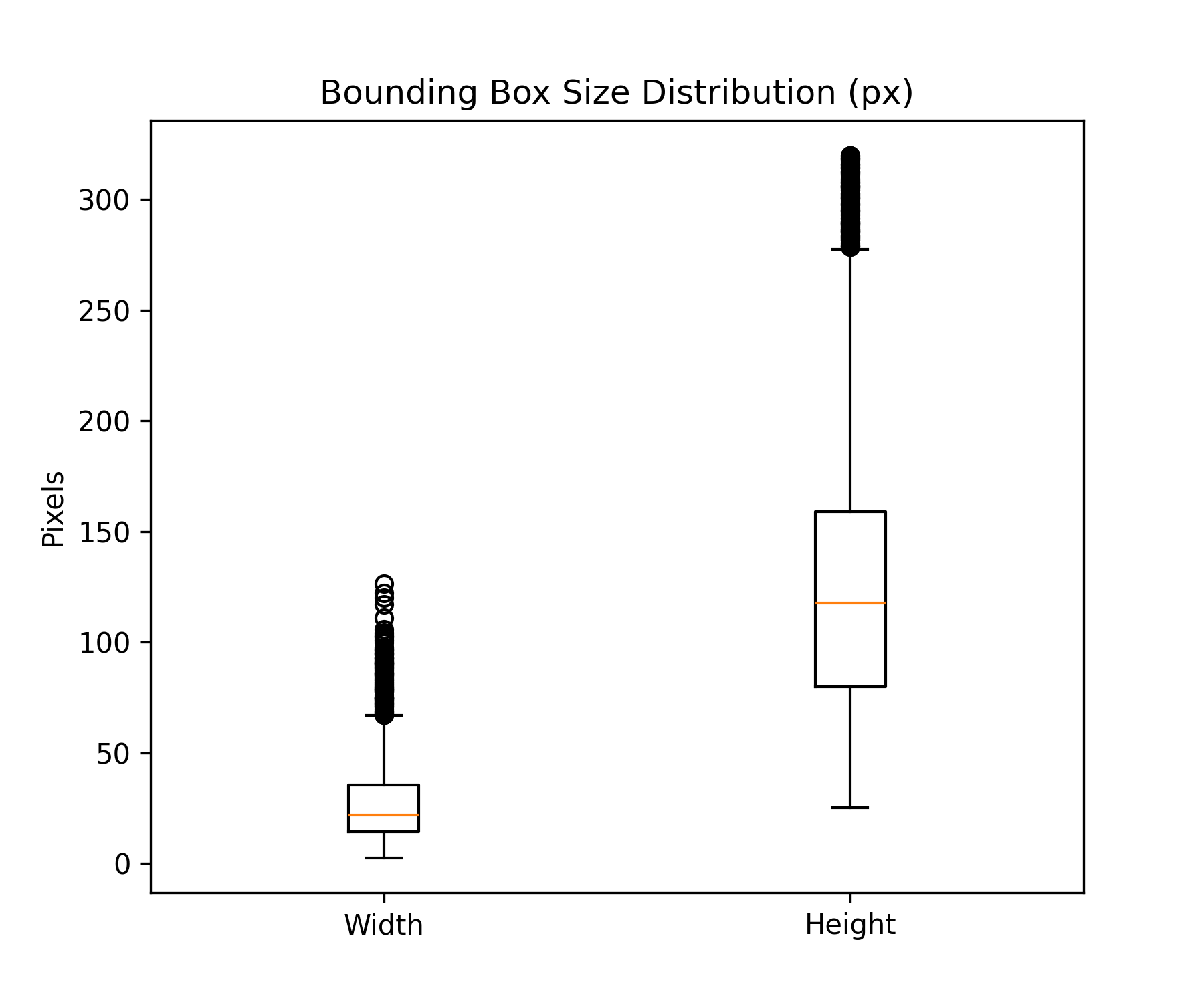}
            \caption{\centering Bounding box size distribution (Width \& Height)}
        \end{subfigure}
    \end{minipage}%
    \hfill
    \begin{minipage}{0.48\linewidth}
        \begin{subfigure}{\linewidth}
            \centering
            \includegraphics[width=\linewidth]{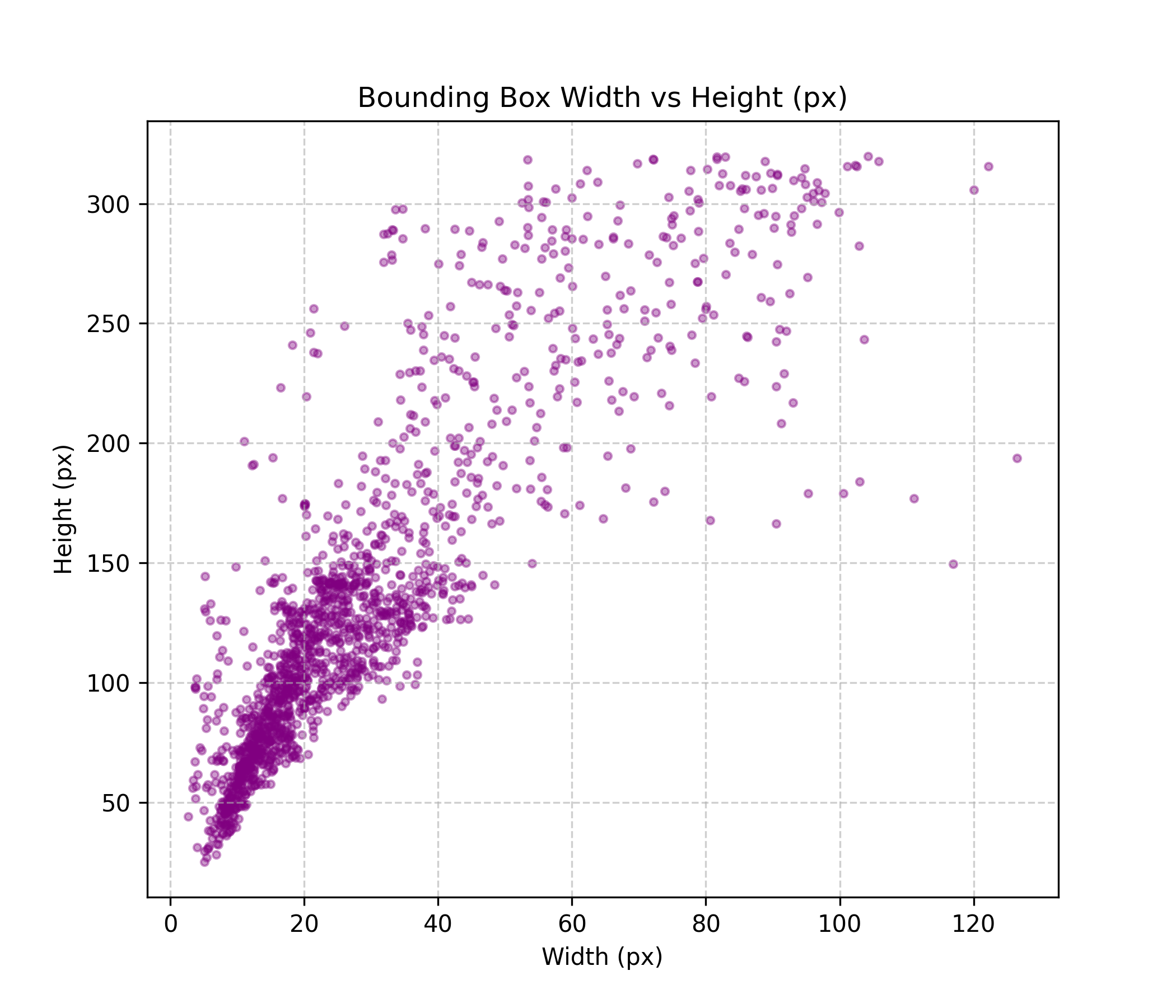}
            \caption{\centering Bounding box width vs height scatter plot}
        \end{subfigure}
    \end{minipage}%
    }
    \caption{Quantitative analysis of YOLO11-4K detection results: (a) Distribution of bounding box sizes, (b) Width--Height relationship highlighting extremely small and large objects.}
    \label{fig:quantitative_results}
\end{figure}

We further analyse the distribution of detected objects across the test set. In Figure~\ref{fig:quantitative_results}, bounding box statistics for YOLO11-4K are presented.  In Figure~\ref{fig:quantitative_results} (a), a boxplot summarising the overall size variability of detected objects is shown, highlighting the prevalence of very small objects, while Figure~\ref{fig:quantitative_results} (b) presents a scatter plot of bounding box width versus height, demonstrating that YOLO11-4K successfully detects extremely small objects (as small as 2.7 × 25.14 pixels(px)). Across the test set, a total of 1,604 objects were detected with an average size of 28.9 × 133.2 px, confirming the model’s strong capability for small object detection in high-resolution 4K panoramic imagery and complementing the qualitative examples of occluded and extremely small object detection.

YOLO11-4K establishes a new state of the art for real-time 4K omnidirectional object detection, outperforming previous YOLO variants and other lightweight architectures. Overall, it provides a practical, high-throughput solution for real-time 360$^{\circ}$ detection, effectively capturing small-scale targets without compromising inference speed. 

\subsection{Cross-dataset generalisation}
To evaluate cross-dataset generalisation, YOLO11-4K was tested on the 4K Multi-Resolution Traffic Monitoring Dataset (MRTMD) \cite{11071529}, which contains multi-class traffic scenes with significant scale variation and many ultra-small objects. Unlike the 360$^{\circ}$ equirectangular imagery in CVIP360, MRTMD consists of standard planar images, making it a challenging domain shift. Nevertheless, the architectural enhancements in YOLO11-4K, particularly the P2 detection head and the hybrid GhostConv/Conv backbone are designed to preserve fine spatial detail and are expected to benefit small-object detection even outside the omnidirectional setting. All MRTMD evaluations were conducted at native 4K resolution using the same thresholds and IoU criteria as in CVIP360.

The cross-model performance of YOLOv5, YOLOv8, YOLO11, and YOLO11-4K on the challenging MRTMD dataset is presented in Table~\ref{MRTMD}. Overall, absolute mAP values are low for all models due to the limited training data, extreme scale variation, dense object arrangements and the prevalence of ultra-small distant objects. Despite these challenges, YOLO11-4K consistently outperformed baseline models, highlighting the effectiveness of its architectural innovations. In particular, the model achieved a mAP of 0.232 on the bicycle class, which predominantly consists of small and distant targets, demonstrating the benefits of the P2 detection head and the hybrid GhostConv/Conv backbone in preserving fine-grained spatial features. While performance on rare classes such as cars exhibited high variance due to limited samples, the overall trend confirms that YOLO11-4K generalises beyond 360$^\circ$ panoramic imagery and improves small-object detection in standard high-resolution scenes, maintaining a balance between accuracy and inference efficiency.

\begin{table}[H]
\caption{Cross-Model Performance Comparison on MRTMD Dataset}
\centering
\begin{tabular}{|p{2cm}|p{2cm}|p{1.5cm}|p{1.5cm}|p{2cm}|p{2cm}|}
\hline
\textbf{Model} & \textbf{Class}  & \textbf{Precision} & \textbf{Recall} & \textbf{mAP50} & \textbf{mAP50-95} \\
\hline
\multirow{5}{*}{YOLOv5} 
& All        & 0.0534  & 0.130  & 0.0293 & 0.0140 \\
& Bicycle    & 0.1760  & 0.0805 & 0.0978 & 0.0417 \\
& Car        & 0.000461 & 0.417 & 0.000346 & 0.000149 \\
& Motorcycle  & 0.00951 & 0.00887 & 0.00480 & 0.00342 \\
& Bus        & 0.0276  & 0.0152 & 0.0141 & 0.0107 \\

\multirow{5}{*}{YOLOv8} 
& All    &   0.0626 & 0.208 & 0.0410 & 0.0201 \\
& Bicycle    & 0.0849 & 0.0211 & 0.0434 & 0.0117 \\
& Car           & 0.000225 & 0.417 & 0.00017 &0.000052 \\
& Motorcycle  & \textbf{0.1600} & \textbf{0.385} & \textbf{0.1180} & \textbf{0.0670} \\
& Bus       & 0.00507 & 0.00891 & 0.00259 & 0.00172 \\

\multirow{5}{*}{YOLO11} 
& All        & 0.0438  & 0.218  & 0.0284 & 0.0171 \\
& Bicycle    &0.0333  & 0.00558 & 0.0168 & 0.00311 \\
& Car         & \textbf{0.000211} & \textbf{0.500} & \textbf{0.000182} & \textbf{0.0000665} \\
& Motorcycle & 0.1360 & 0.360 & 0.0937 & 0.0634 \\
& Bus        & \textbf{0.00556} & \textbf{0.00629} & \textbf{0.00282} & \textbf{0.00194} \\

\multirow{5}{*}{\textbf{YOLO11-4K}} 
& All        & \textbf{0.117}  & 0.0761 & \textbf{0.0600} & \textbf{0.0352} \\
& Bicycle    & \textbf{0.452}  & \textbf{0.0223} & \textbf{0.2320} & \textbf{0.1350} \\
& Car           & 0.000155 & 0.250 & 9.97e-05 & 5.35e-05 \\
& Motorcycle  & 0.0151 & 0.0315 & 0.00777 & 0.00527 \\
& Bus         & 0.000844 & 0.000524 & 0.000426 & 0.000299 \\
\hline

\end{tabular}
\label{MRTMD}
\end{table}

\section{Ablation Studies}
\label{sec:ablation}

To understand the contribution of each architectural component in YOLO11-4K, we conducted systematic ablation experiments using 3840×3840 resolution inputs. Each ablation configuration was evaluated under a 5-fold cross-validation protocol, and results were averaged across folds to reduce variance and better reflect model stability, also expressed in Table~\ref{ablation}. The baseline model YOLO11s achieved a strong mAP@50 of 0.908 with an inference time of 112.3 ms.

\begin{table}[htbp]
\caption{Ablation Study of YOLO11s Variants on 4K Input}
\centering
\resizebox{\linewidth}{!}{%
\begin{tabular}{|l|l|l|l|l|}
\hline
\textbf{Model} & \textbf{Parameters} & \textbf{GFLOPs} & \textbf{mAP @50} & \textbf{Inference (ms)} \\
\hline
YOLO11s (Baseline) & 2,582,542 & 6.3 & 0.904±0.011 & 112.3±0.15\\

YOLO11s+P2 Head & 2,634,248 & 10.1 & 0.908±0.002 & 126.6±0.12 \\

YOLO11s+P2+ LightweightBB & 4,846,512 & 13.9 & 0.867±0.023 & 102.6±0.12 \\

–GhostConv All & 4,224,896 & 12.3 & 0.873±0.012 & 104.46±0.162 \\

–Hybrid (GhostConv+Conv) & 1,149,212 & 7.0 & 0.888±0.021 & 71.2±0.126 \\

–Hybrid+ C3Ghost & 786,662 & 5.2 & 0.847±0.022 & 61.38±0.098 \\

\textbf{-Hybrid+ kernel–stride parameter tuning }& \textbf{1,377,444} & \textbf{2.4} & \textbf{0.950±0.007} & \textbf{28.3±0.126}\\
\hline
\end{tabular}
}

\label{ablation}

\end{table}

Introducing a P2 detection head in the YOLO11-baseline provided a slight gain in detection accuracy by enabling the model to better capture fine-grained features from early layers. This enhancement came at the cost of a modest increase in parameters and latency.

Next, we tested a variant lightweightBB (lightweight backbone) where the standard backbone was replaced by a computationally optimised architecture. This lightweight backbone reduces complexity through gradual spatial down-sampling using strided convolutions, compact C3k2 modules with fewer channels in early layers and 1×1 convolutions to compress feature maps where possible. While this configuration reduced inference time to 102.6 ms, it also caused a drop in accuracy (mAP@50: 0.867), indicating a trade-off between speed and precision.

Replacing all standard convolutions with GhostConv modules led to further efficiency gains, reducing both parameter count and computational load while maintaining a comparable mAP@50 of 0.873, though inference slowed slightly to 104.46ms. The inference time increases as GhostConv operations are less optimised for GPU parallelisation, causing extra kernel launches and memory overhead. Although GhostConv reduces FLOPs, its fragmented operations slow the execution compared to standard Conv layers, thus using GhostConv selectively can help regain speed efficiency.

We also explored a hybrid model that blends GhostConv in shallow layers and standard convolution in deeper layers. This setup offers one of the better balances between accuracy and speed, achieving mAP@50 of 0.888 with an inference time of just 71.2 ms. The increased speed of the hybrid model compared to the full GhostConv model can be attributed to the differing computational efficiencies of convolution types across network depths. GhostConv is particularly effective in early, shallow layers with large feature maps by minimising redundancy and computation. In contrast, deeper layers that have smaller spatial dimensions but more channels may experience overheads from GhostConv’s linear operations and concatenation steps, thereby reducing its efficiency. By employing standard convolutions in these deeper layers, the hybrid model attains faster inference. This strategy capitalises on GhostConv’s efficiency in initial layers and the speed advantages of standard convolutions deeper in the network, resulting in a more favourable balance between accuracy and performance.

The compact version using both GhostConv and C3Ghost (a lightweight version of the YOLO C3 module that replaces standard convolutions with Ghost Convolutions to reduce computations and model size while maintaining similar accuracy)\cite{xiao2025c3ghost} modules resulted in the fewest parameters (786K) and fast inference (61.38 ms). However, it also recorded the lowest mAP@50 (0.847), suggesting a significant trade-off for extreme efficiency.

Finally, the Hybrid model with kernel–stride parameter tuning (adjusting kernel sizes and strides to reduce feature map dimensions and improve efficiency) achieved a strong balance between accuracy and speed. Compared to the baseline YOLO11s (2.58M parameters, 6.3 GFLOPs, 0.917 mAP@50, 112.3 ms inference), this variant required only 1.38M parameters and 2.4 GFLOPs, yet delivered the highest mAP@50 of 0.95. Moreover, it achieved an inference time of just 28.3 ms which is much faster than the baseline, demonstrating that careful kernel–stride adjustments can significantly reduce computation while simultaneously boosting detection performance.

Overall, these experiments demonstrate how each modification affects the speed–accuracy trade-off, guiding the design of efficient models for high-resolution small-object detection in panoramic imagery. While YOLO11-4K is applicable to high-resolution images in general, it is specifically fine-tuned on panoramic 360$^{\circ}$ images, enhancing robustness to spatial distortions, wide fields of view, and non-uniform scaling inherent in equirectangular projections. This fine-tuning improves detection accuracy for small and distorted objects commonly encountered in panoramic imagery.

\section{Conclusion}
\label{sec:conclusion}
In this work, we presented YOLO11-4K, a real-time detection framework designed for high-resolution 4K panoramic imagery. By integrating a lightweight GhostConv backbone and a dedicated P2 detection layer, the model improves small-object sensitivity while significantly reducing computational cost. Experiments on the CVIP360 dataset show that YOLO11-4K achieves substantial speed improvements, reducing inference time by nearly 75\% while maintaining strong detection performance, demonstrating an effective balance between accuracy and efficiency for 360$^{\circ}$ environments, which remain particularly challenging due to extreme distortion and the need for high resolution processing. The results further indicate robust generalisation across diverse panoramic scenes. While this study focused on 4K inputs, the architectural principles introduced here are broadly applicable to other high-resolution domains, including autonomous navigation, large-scale surveillance and immersive AR/VR systems. Future work will explore adaptive spherical representations and integration with multi-object tracking pipelines for full real-time 360$^{\circ}$ perception.

\section{Limitations and Future Work}
\label{sec:limitations}
A key limitation of this study is the scarcity of large-scale, publicly available annotated 4K omnidirectional datasets. To the best of our knowledge, CVIP360 remains the only dataset offering native 4K resolution with bounding box annotations suitable for small-object detection. While this work establishes strong baselines and demonstrates the advantages of the proposed YOLO11-4K architecture on CVIP360, further validation on additional high-resolution datasets would strengthen generalisability claims. As part of future work, we plan to: (i) extend experiments to synthetic or semi-automatically annotated 4K panoramas to broaden evaluation coverage; and (ii) explore cross-resolution strategies, including domain adaptation from lower-resolution panoramic datasets (e.g., 1080p, 2K) and scaling analysis up to 4K, to quantify robustness. Addressing these challenges will help establish a comprehensive benchmark for real-time small object detection in ultra-high-resolution omnidirectional imagery. Although YOLO11-4K demonstrates improved small-object sensitivity on MRTMD, absolute performance remains low due to extreme scale variation, highlighting the need for high-resolution multi-class benchmarks.

\section*{Acknowledgements}
\label{sec:acknowledgements}

This work was supported by the \textbf{National Computational Infrastructure (NCI)}, which provided the high-performance computing resources used for training and evaluation. The first author also acknowledges support from a UNSW PhD Scholarship.

\end{document}